%% file: main.tex
\documentclass[11pt]{article}

\usepackage[preprint]{acl}

\usepackage{times}
\usepackage{latexsym}
\usepackage{hyperref}
\usepackage{subcaption}
\usepackage[T1]{fontenc}

\usepackage[utf8]{inputenc}

\usepackage{booktabs}
\usepackage{multirow}
\usepackage{siunitx}
\usepackage{makecell}
\usepackage{amsmath}

\usepackage{microtype}

\usepackage{inconsolata}

\usepackage{graphicx}
\usepackage{pgfplots}
\usepgfplotslibrary{groupplots}
\pgfplotsset{compat=1.18}

\definecolor{codegreen}{rgb}{0,0.6,0}
\definecolor{codegray}{rgb}{0.5,0.5,0.5}
\definecolor{codepurple}{rgb}{0.58,0,0.82}
\definecolor{backcolour}{rgb}{0.95,0.95,0.92}
\definecolor{codered}{rgb}{0.8,0,0}
\usepackage{listings}%
\title{Symbolic Grounding Reveals Representational Bottlenecks in Abstract Visual Reasoning}

\author{%
\textbf{Mohit Vaishnav}\textsuperscript{\textnormal{1,2}}~~~~ 
\textbf{Tanel Tammet}\textsuperscript{\textnormal{1}} \\
   \textsuperscript{1} Applied Artificial Intelligence Group, Tallinn University of Technology, Estonia 12169\\
   \textsuperscript{2} Kimova AI, Tallinn, Estonia 10113\\
  \texttt{mohit.vaishnav@taltech.ee}
}

\begin{document}
\maketitle

\begin{abstract}

Vision--language models (VLMs) often fail on abstract visual reasoning benchmarks such as Bongard problems, raising the question of whether the main bottleneck lies in reasoning or representation. We study this on Bongard-LOGO, a synthetic benchmark of abstract concept learning with ground-truth generative programs, by comparing end-to-end VLMs on raw images with large language models (LLMs) given symbolic inputs derived from those images. Using symbolic inputs as a diagnostic probe rather than a practical multimodal architecture, our \emph{Componential--Grammatical (C--G)} paradigm reformulates Bongard-LOGO as a symbolic reasoning task based on LOGO-style action programs or structured descriptions. LLMs achieve large and consistent gains, reaching mid--90s accuracy on Free-form problems, while a strong visual baseline remains near chance under matched task definitions. Ablations on input format, explicit concept prompts, and minimal visual grounding show that these factors matter much less than the shift from pixels to symbolic structure. These results identify representation as a key bottleneck in abstract visual reasoning and show how symbolic input can serve as a controlled diagnostic upper bound.

\end{abstract}

\section{Introduction}

Vision--language models (VLMs) have achieved strong results on captioning, retrieval, and recognition from large-scale paired data \citep{radford2021clip, jia2021align, ramachandran2025gpt4o}, yet they often fail on tasks that require abstract reasoning, compositional generalization, or spatial inference \citep{yuksekgonul2022behave, parascandolo2025causal, li2025enhancing_compositional}. Analyses of CLIP-style encoders suggest that end-to-end architectures may overemphasize salient objects or early caption tokens while underrepresenting relational and structural information \citep{fine2025bias_clip, kang2023cooccurrence, campbell2024binding}. This raises a basic question: is the main bottleneck in current multimodal models their \emph{reasoning} mechanisms, or the \emph{representations} that feed those mechanisms?

\begin{figure}[t]
    \centering
    \includegraphics[width=\linewidth]{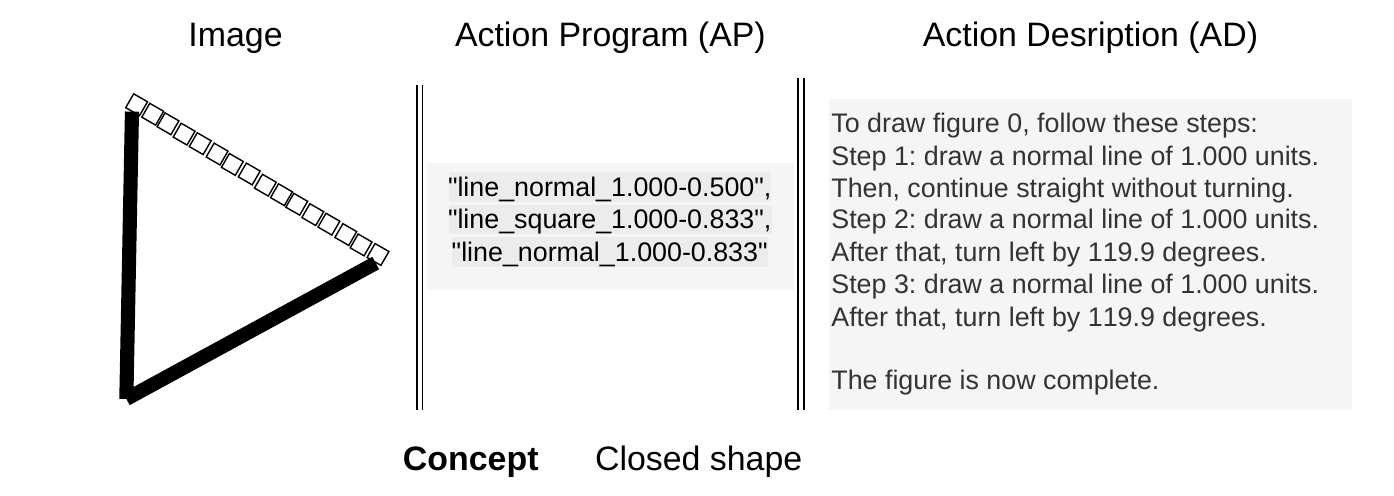}
    \caption{
    Example Bongard-LOGO instance under the Componential--Grammatical interface.
    From left to right we show the rendered shape, its action-program (AP) representation, and its action-description (AD) rendering.
    In the concept-conditioned variant, the prompt additionally provides the high-level concept label, allowing us to compare visual, procedural, and natural-language interfaces for the same underlying example.
    }
    \label{fig:logo_example}
\end{figure}

We address this question in a controlled synthetic setting: \textbf{Bongard-LOGO} \citep{nie2020bongardlogo}, a benchmark of concept-learning tasks inspired by classic Bongard problems \citep{bongard1970pattern, foundalis2006phaeaco}. Each problem presents positive and negative example images and requires inferring a latent rule that separates them. Figure~\ref{fig:logo_example} previews the three interfaces used in our experiments: the rendered image itself, its procedural action program, and its natural-language action description. Unlike conventional benchmarks, Bongard-LOGO emphasizes geometric, relational and topological properties--symmetry, convexity, connectivity--rather than object semantics, and exposes the ground-truth procedural LOGO program that generates each image.

Classic work on Bongard problems argues that solving them requires a \emph{visual language} of structured primitives and relations; Bayesian and causal neuro-symbolic systems define such vocabularies and inference procedures, achieving strong performance on hand-designed sets while remaining interpretable \citep{depeweg2024visual_language, bongard_architecture_2023}. We follow this line but replace bespoke engines with general-purpose language models as the reasoning backend.

We introduce the \textbf{Componential--Grammatical (C--G)} paradigm: a modular interface that decouples perception from reasoning by posing Bongard-LOGO as a symbolic reasoning problem. Instead of operating on pixels, models receive the underlying \emph{action programs} that generated the images or structured textual descriptions derived from them, forming a LOGO-style visual language. A language model then induces a rule and classifies query examples from symbolic input alone. This setup lets us ask: to what extent do recent language models exhibit abstract generalization when supplied with explicit symbolic structure, and how do representation format, concept conditioning, and minimal visual grounding modulate their performance?

Empirically, a strong VLM baseline (Gemini-2.5-Flash) performs near chance ($\sim$50\%) on Bongard-LOGO, whereas the same tasks reframed via symbolic input yield substantially higher accuracies (often 60--80\% across models, with Phi-4-Reasoning reaching 96.2\% on Free-form problems). The main driver of this gap is representational: compositional, interpretable structure enables generalization, while variations in format, concept prompts, and grounding yield smaller, model-dependent effects. Although symbolic input is not a realistic perceptual modality, we treat C--G as a diagnostic upper bound on what current models can do once grounded in an appropriate visual language, rather than as a complete solution to perception.

Our goal is diagnostic rather than architectural. We do not introduce a new end-to-end multimodal system or learned visual front-end; instead, we vary the \emph{representational interface} between perception and reasoning on Bongard-LOGO. By comparing the same family of language models under procedural programs, natural-language descriptions, concept prompts, and structure-breaking perturbations, we isolate how input format and inductive bias shape abstract generalization independently of vision encoder choice.

Concretely, our contributions are threefold.
\textit{First}, we present the Componential--Grammatical (C--G) interface for Bongard-LOGO and show that representational format alone can induce 25--30 point gains over a strong visual baseline across 12 models.
\textit{Second}, we systematically probe this interface via concept conditioning, minimal-context prompting, grounded variants, and token randomization controls, revealing when and how symbolic structure—rather than surface statistics—drives success.
\textit{Third}, we provide a cross-model analysis of capacity and brittleness under these interfaces, positioning C--G as a diagnostic tool for studying representational bottlenecks in multimodal reasoning.

While prior work has shown that specialized symbolic solvers perform well on Bongard problems, it is not obvious that general-purpose pretrained LLMs can induce abstraction rules directly from procedural symbolic traces without domain-specific search or operators. Our experiments show that once grounded in an explicit visual language, modern LLMs can function as effective abstract inference engines.

\section{Related Work}

\paragraph{Vision--Language Models and Grounding.}
Modern VLMs such as CLIP and ALIGN learn joint image--text representations and excel at recognition and retrieval \citep{radford2021clip, jia2021align, ramachandran2025gpt4o}, but benchmarks targeting compositional generalization, spatial reasoning, and relational understanding reveal systematic failures \citep{yuksekgonul2022behave, ma2023crepe, hu2023tifa, hsieh2023sugarcrepe}. These models often prioritize coarse object or token statistics, with limited sensitivity to attribute binding, part--whole structure, or word order, and analyses trace this to architectural and training biases such as overreliance on early caption tokens or spatial salience \citep{fine2025bias_clip, kang2023cooccurrence, parascandolo2025causal, campbell2024binding}. Recent work introduces more explicit intermediate structure, via multi-granular alignment between sentence parts and image regions \citep{le2025progressive}, generative-flow-style chain-of-thought (CoT) reasoning over images \citep{kang2025gflowvlm}, or program-like plans over symbolic state \citep{xu2025vlagent}. In contrast, we change only the \emph{input format}—from pixels to LOGO-style symbolic programs—and ask how this affects abstract generalization in language models.

\paragraph{Bongard Problems and Visual Abstraction.}
Bongard problems are classic visual reasoning tasks where a rule must be induced to separate positive from negative examples \citep{bongard1970pattern, hofstadter1979geb, foundalis2006phaeaco}. They emphasize geometric and topological concepts such as symmetry, containment, and convexity, and have informed work on analogy and visual concept learning \citep{gentner1997analogy, gentner2017analogy}. Bongard-LOGO \citep{nie2020bongardlogo} extends this paradigm into a synthetic benchmark of 12{,}000 procedurally generated problems from ground-truth LOGO programs. Despite their visual simplicity, these tasks are difficult for current models: even advanced systems like GPT-4o and Gemini attain only 50--60\% accuracy on abstract Bongard-style sets, far below human performance \citep{wust2025bongard}. Several systems treat Bongard problems as inference in an explicit visual language: \citet{depeweg2024visual_language} define a symbolic vocabulary and use Bayesian inference with pragmatic constraints, and Bongard Architecture implements a causal, operator-based neuro-symbolic system that solves nearly all Maksimov--Bongard puzzles while remaining interpretable \citep{bongard_architecture_2023}. Bongard-in-Wonderland shows that frontier foundation models still often fail on simple geometric concepts like spirals \citep{wust2025bongard}. Our work follows this tradition in using a structured visual language, but replaces bespoke inference engines with general-purpose LLMs and leverages Bongard-LOGO's generative programs as the underlying representation.

\paragraph{Symbolic Interfaces and Neuro-Symbolic Models.}
Our approach fits within a broader effort to modularize perception and reasoning through structured interfaces. Componential Analysis (CA) \citep{vaishnav2025cognitive} uses a VLM to generate natural-language scene descriptions that an LLM then uses for classification or inference; this works well on natural-image Bongard datasets such as Bongard-OpenWorld \citep{wu2024bongardopenworld}, but performs poorly on Bongard-LOGO, where precise relations are hard to verbalize \citep{peng2024spec}. Inverse-graphics and shape-grammar approaches similarly emphasize program-like representations of visual structure \citep{stiny1972shape, gips1975shape, knight2015making, kulkarni2015dcign, yildirim2020efficient, kulits2024inverse}. For Bongard-LOGO, probabilistic concept models combined with Sinkhorn distances over symbolic shapes \citep{sbss_pmoC_2024} and ARC-style neuro-symbolic agents that search in task-specific languages \citep{nsa_arc_2025} illustrate the value of explicit languages as grounding interfaces. Our Componential--Grammatical (C--G) interface instantiates this pattern for LOGO-based Bongard problems with LLMs as the reasoning backend and allows us to evaluate how procedural programs versus descriptive text modulate downstream reasoning performance independently of perceptual noise.

\section{Method}

\paragraph{Bongard-LOGO as a Diagnostic Benchmark.}
We adopt the LOGO benchmark \citep{nie2020bongardlogo} as a controlled testbed for studying the interface between perception and reasoning. Each of its 12{,}000 problems consists of six positive and six negative example images, along with a query image to classify. All images are procedurally generated from known LOGO programs, enabling precise control over visual properties. Following the released benchmark splits, we report results for \emph{Free-form} (FF), \emph{Basic} (BD), and two \emph{Human-designed} evaluation splits: HD-Comb and HD-Novel. FF concepts are defined by stroke sequences, BD concepts by reusable shape categories and their combinations, and the two HD splits test transfer to human-designed shapes under combinatorial and novel-shape regimes. Unless otherwise noted, HD denotes the mean of HD-Comb and HD-Novel.

\paragraph{Symbolic Representation via Procedural Grammars.}
Unlike typical image benchmarks, Bongard-LOGO provides access to the full generative trace for each image. Each figure is constructed from a hierarchy of LOGO-like drawing instructions: a \texttt{BongardImage} comprises one or more \texttt{OneStrokeShape} objects, each built from \texttt{BasicAction} primitives (e.g., \texttt{Line}, \texttt{Arc}). We encode these structures as compact grammar strings that preserve the sequence of operations, angle information, and shape identifiers. For instance, a triangle might be represented as a sequence of three equal-length line segments with 120-degree turns between them. These symbolic traces form a procedural shape grammar that is both noise-free and compositional, aligning with inverse-graphics views of visual processing \citep{stiny1972shape, knight2015making, kulkarni2015dcign, kulits2024inverse}.

\paragraph{Componential–Grammatical Paradigm (C–G).}
We introduce the \textbf{Componential–Grammatical (C–G)} paradigm, a modular pipeline that separates visual structure recovery from downstream reasoning. C–G operates in two stages:

\begin{itemize}
    \item \textbf{Symbolic Grounding:} Each image in a Bongard-LOGO problem (positives, negatives, and query) is represented as a symbolic input—either a procedural grammar string or a descriptive natural-language version—using its ground-truth program.
    \item \textbf{Textual Reasoning:} A language model receives only the symbolic inputs and must infer a rule and classify the query as positive or negative. No pixel data or visual features are provided.
\end{itemize}

This setup transforms the task from multimodal perception-plus-reasoning to a purely symbolic inference problem. Crucially, it allows us to isolate the role of input representation in driving model behavior. If LLMs perform well in the symbolic setting but VLMs fail on raw images, this pattern points to a representational or grounding bottleneck rather than a lack of reasoning capacity.

\begin{figure*}[t]
    \centering
    \includegraphics[width=.85\linewidth]{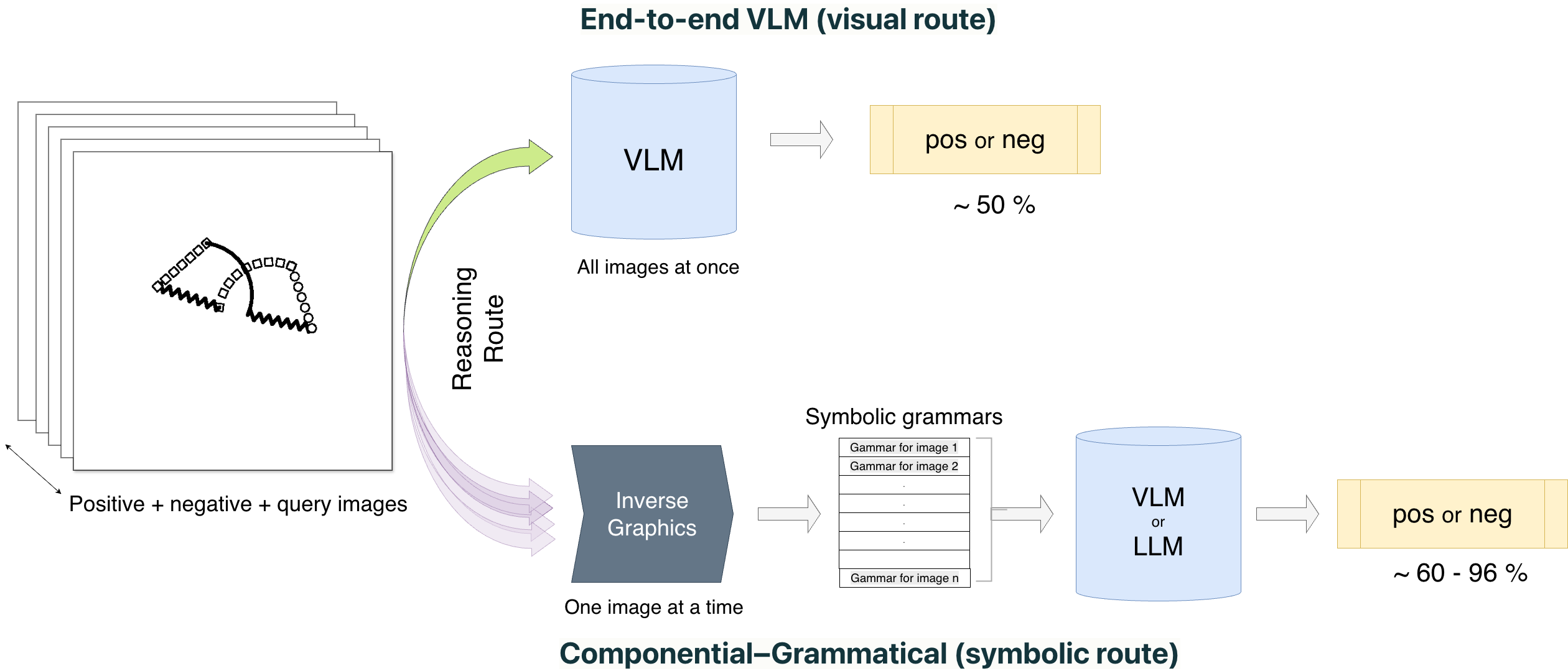}
    \caption{
    Comparison of visual and symbolic pipelines on Bongard-LOGO. 
    In the \emph{visual route} (top), a Vision--Language Model (VLM) receives images directly and must infer both structure and rules, typically yielding near-chance accuracy. 
    In the \emph{symbolic route} (bottom), problem instances are represented by LOGO-style action programs; a language model then performs rule induction and classification using these symbolic representations, achieving substantially higher performance on the same task.
    }
    \label{fig:cg_overview}
\end{figure*}

\section{Experiments}
\label{sec:experiments}
We design our experiments to answer three questions: (i) do symbolic representations unlock abstract visual reasoning that VLMs fail to exhibit on Bongard-LOGO, (ii) how does the \emph{form} of symbolic input (programs vs.\ descriptions, with or without concepts and images) modulate performance, and (iii) do models rely on \emph{rule-like structure} rather than exploiting token-frequency or lexical-overlap heuristics?

\paragraph{Models.}
We evaluate proprietary Gemini models \citep{google2024gemini}, using Gemini-2.5-Flash as the primary end-to-end visual baseline and including Gemini-3-Flash in the supplementary model-capacity analysis. We also evaluate open-source models including DeepSeek-R1 (8B, 14B, 32B) \citep{guo2025deepseekr1}, Phi-4 and Phi-4-Reasoning (14B) \citep{microsoft2025phi4}, Qwen2.5 and Qwen3 (up to 32B) \citep{yang2025qwen3}, Gemma 3 (12B, 27B), and Mistral Magistral. Open models are run locally using the Ollama framework. Refer to Appendix~\ref{app:models} for more details.

\paragraph{Evaluation Protocol.}
Each Bongard-LOGO problem contains 6 positive examples, 6 negative examples, and 1 query image (detailed in Appendix~\ref{app:example_instance}). Models must predict whether the query belongs to the positive or negative class. Accuracy is computed over a fixed 2{,}000-problem subset containing 500 problems from each reported evaluation split: BD, FF, HD-Comb, and HD-Novel. The subset is drawn uniformly at random from the test split without any performance-based filtering, and all models are evaluated on the same subset for comparability. Open-source models use deterministic decoding (temperature 0.0, no sampling); Gemini models use low-temperature official modes. Outputs are parsed from a constrained JSON format containing a reasoning trace, induced rule, and binary label; unparsable outputs are counted as errors. Exact prompts, model identifiers, and infrastructure details appear in Appendix~\ref{app:reproducibility},\ref{app:prompts}.

\subsection{Main Conditions: From Pixels to Symbols}

Our primary comparison contrasts an end-to-end visual baseline with variants of the Componential--Grammatical (C--G) interface, where models operate on symbolic input only.

\paragraph{Visual VLM Baseline.}
Gemini-2.5-Flash receives all 13 \emph{images} and directly predicts the query label. This represents a standard end-to-end VLM pipeline in which perception and reasoning are tightly coupled.

\paragraph{C--G (Formal Grammar).}
Models receive only procedural shape grammar strings (as shown in Appendix~\ref{app:eg_action_program}) derived from the ground-truth LOGO programs for all 13 images, with no pixel input. This condition tests whether explicit, compositional structure alone enables high-accuracy rule induction and classification.

\paragraph{C--G (Natural Language).}
Instead of grammar strings, models see step-by-step English descriptions of the drawing actions and geometric relations as seen in Appendix~\ref{app:eg_action_description}. This probes whether recasting the same structure in familiar linguistic form helps or harms abstract reasoning relative to a compact formal syntax.

\subsection{Ablations: How Much Structure and Supervision?}

We next vary the amount and form of guidance around the symbolic input.

\paragraph{Minimal-Context Grammar.}
Models receive raw grammar strings plus a brief instruction describing the task, but no explanation of the formalism itself. This few-shot setting asks whether models can infer a novel syntactic interface and the intended rule-induction task from examples alone.

\paragraph{Concept-Conditioned C--G.}
The high-level target concept (e.g., ``is convex'', ``contains a triangle'' Appendix~\ref{app:eg_concept}) is provided in the system prompt alongside the action programs. This converts the task from rule discovery to rule verification and tests whether explicit concept supervision provides additional gains over purely example-based learning. 

\paragraph{Grounded C--G.}
Models receive symbolic programs for all 13 images plus the \emph{rendered query image}. This condition probes whether a single visual anchor, tied to otherwise symbolic input, yields further improvements beyond the purely symbolic setting. Because it requires image input, we restrict this variant to VLMs that accept both text and images (Gemma3:12B, Gemma3:27B, LLaVA-Llama3, and Qwen2.5-VL:32B).

\subsection{Structure vs.\ Surface: Token Randomization Tests}

Finally, we test whether models rely on genuine rule learning over symbolic structure rather than superficial token matching.

\paragraph{Randomizing Query Programs.}
In the first control, we randomly permute the sequence of drawing actions in the \emph{query} program while keeping the support examples, the query image, and its label unchanged. If models primarily exploit token inventory alone, performance should remain similar; if they rely on structured comparisons between query and support examples, accuracy should degrade once the ordered composition of the query is broken.

\paragraph{Shuffling Positive and Negative Sets.}
In the second control, we randomly reassign the 12 support examples between the positive and negative sets while leaving the query program, query image, and query label untouched. This preserves within-problem token inventories but destroys the original support-set partition. Performance therefore need not fall exactly to 50\%: after shuffling, the support sets can still instantiate alternative regularities that partially align with the unchanged query. The diagnostic signal is the drop relative to the unperturbed Action Program condition.

We report these controls only for BD and FF. HD performance is already comparatively low in the unperturbed setting, so the same perturbations would be less informative there.

Together, these perturbation studies test whether high C--G performance depends on relational and compositional structure in the symbolic input, rather than on simple token-frequency heuristics alone.

\section{Results}
\label{sec:results}

We report findings from systematic evaluation of the Componential–Grammatical (C–G) paradigm across 12 models and multiple input regimes. Our core result is consistent: when models receive structured symbolic input in place of raw perceptual input, performance improves dramatically, especially on conceptually grounded reasoning tasks.

\subsection{Main Effects of Symbolic Grounding}
\label{sec:results_main}

Table~\ref{tab:main_results} reports mean accuracy across three benchmark categories and key experimental conditions. Gemini-2.5-Flash achieves chance-level performance (50–51\%) across the board when operating on raw images. In contrast, when provided with symbolic input (Action Programs), models reach up to 96.2\% accuracy (Phi-4-Reasoning, Free-form), with average gains of roughly 20–30 points on BD and FF and smaller, but still substantial, gains on HD over the visual baseline.

\begin{table}[t]
\centering
\small
\caption{
Summary of key results across experimental conditions on Bongard-LOGO.
The first five rows and the randomization rows are mean accuracy (\%) across 12 language models.
Rows marked with \textsuperscript{$\dagger$} are averaged over the matched 4-model VLM-capable subset (Gemma3 12B/27B, LLaVA-Llama3, Qwen2.5-VL 32B) and should be compared only within that subset.
`Base' denotes the minimal-context symbolic baseline.
AP = Action Program (formal LOGO-style grammar), AD = Action Description (natural-language description of actions).
}
\label{tab:main_results}
\begin{tabular}{lccc}
\toprule
\textbf{Condition} & \textbf{FF} & \textbf{BD} & \textbf{HD} \\
\midrule
Visual VLM (Gemini-2.5)          & 50.2 & 49.8 & 50.1 \\
C--G (AP) & 78.1 & 68.8 & 61.0 \\
C--G (AD) & 79.3 & \textbf{72.0} & 59.1 \\
C--G + Concept (AP + C) & \textbf{79.3} & 70.0 & \textbf{61.1} \\
Minimal-Context Grammar & 77.3 & 67.7 & 59.6 \\
\midrule
Visual baseline (VLM subset)\textsuperscript{$\dagger$} & 62.6 & 57.1 & 57.5 \\
Grounded C--G (AP)\textsuperscript{$\dagger$} & 62.4 & 55.3 & 58.2 \\
Grounded C--G (AD)\textsuperscript{$\dagger$} & 61.2 & 55.3 & 55.8 \\
\midrule
Category Permutation & 70.6 & 60.1 & -- \\
Sequence Permutation & 58.0 & 64.2 & -- \\
\bottomrule
\end{tabular}
\vspace{-.5cm}
\end{table}

\begin{figure*}[t]
\centering
\input{per_model_condition_plot.tex}
\caption{
Per-model percentage-point change under symbolic interventions, aggregated from Table~\ref{tab:bongard_results}.
Each gray dot is one model; black diamonds mark mean deltas.
`AD$-$Base' and `AP$-$Base' measure gains over the minimal-context symbolic baseline, while `+C on AP' isolates the incremental change from AP to AP+C.
The zero line marks no change.
This makes the reviewer-facing pattern explicit: Action Descriptions help most on FF and BD, AP is slightly more stable on HD, and concept conditioning yields modest but heterogeneous model-level effects.
}
\label{fig:per_model_symbolic}
\end{figure*}
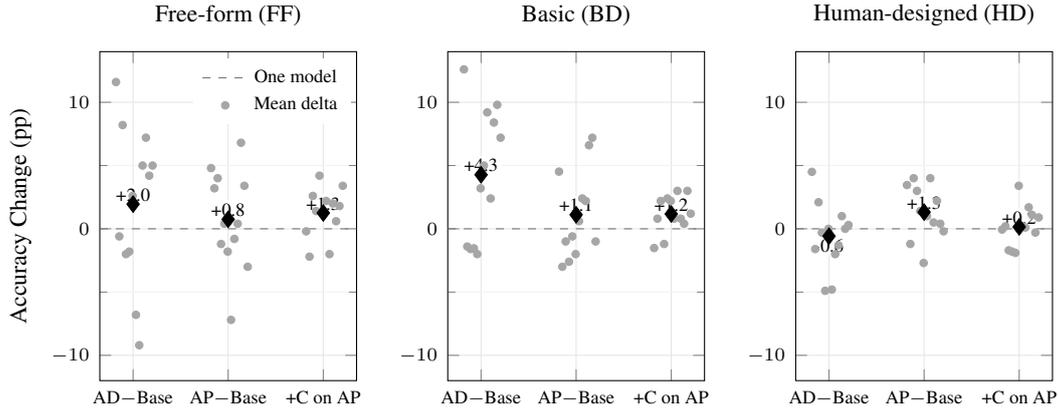


These results confirm that models' poor performance in the visual setting stems primarily from representational limitations rather than an inability to reason over structured input. The pooled symbolic conditions improve markedly over the visual-only Gemini baseline, and the matched four-model subset shows that adding a single query image does not recover additional gains once the same models already receive symbolic structure. Across conditions, FF is easiest to recover from symbolic input, BD is intermediate, and HD remains the most challenging.

Beyond overall accuracy, we also examine how models treat positive vs.\ negative examples under the AP interface.
Across datasets, accuracy on positives substantially exceeds accuracy on negatives, and a chi-square test confirms a strong dependence of correctness on example class (see Appendix~\ref{app:class_asymmetry} for details).
This indicates that models adopt a liberal decision criterion that favors ``positive'' classifications: they are more reliable at recognizing configurations that satisfy a learned rule than at rejecting heterogeneous counterexamples.

\subsection{Descriptive vs. Procedural Input}
\label{sec:results_representation}

When symbolic structure is rendered in natural language, pooled accuracy is higher than AP on FF and BD, while AP retains a modest advantage on HD. Figure~\ref{fig:per_model_symbolic} makes the direction of that difference more explicit: relative to the minimal-context symbolic baseline, AD yields larger mean gains on FF and BD (+2.0 and +4.3 points), whereas on HD the mean AD shift is slightly negative (-0.6) and AP remains modestly positive (+1.3). This pattern is not driven by a single outlier model: the point clouds show that AD is broadly competitive on the easier FF and BD regimes, whereas AP is slightly more stable on the human-designed regime (Table~\ref{tab:main_results}).

\subsection{Concept Conditioning and Minimal Guidance}
Providing models with the target rule label (AP + C) yields modest but heterogeneous effects. Relative to AP alone, the mean change is +1.3 points on FF, +1.2 on BD, and near zero on HD (+0.2), but Figure~\ref{fig:per_model_symbolic} shows that these averages mask mixed model-level behavior rather than a universal improvement. The appendix tables clarify why: some larger models show positive average gains under concept conditioning, whereas smaller or more instruction-sensitive models exhibit mixed or even negative deltas (Tables~\ref{tab:model_size_effect} and~\ref{tab:size_averaged_gains}). We therefore treat AP + C as a model-dependent prompt variant rather than a uniformly stronger condition. The same pattern appears in the minimal-context setting: a shorter prompt can outperform the full prompt for some models, suggesting that additional guidance sometimes introduces distraction or prompt overload rather than useful supervision.

\subsection{Visual Grounding Offers Little Added Benefit}
Within the matched VLM-capable subset, adding a single visual anchor in the Grounded C--G setting leaves accuracy essentially unchanged relative to the corresponding symbolic-only baseline and slightly below the same subset's visual baseline on FF and BD (Table~\ref{tab:main_results}; see Appendix~\ref{app:grounded_cg}). The main result is therefore not that query grounding helps a little, but that once symbolic structure is already available, a single additional image does not unlock further gains.


\subsection{Randomization Controls: Sensitivity to Structure}

To test whether models rely on structured rule induction rather than token-level heuristics, we introduce two perturbations in the Action Program regime. 

In \emph{Category Permutation}, we randomly reassign support examples between positive and negative sets within each problem, while keeping the query and its gold label fixed. This manipulation preserves token inventories but breaks the original support partition. Performance remains above 50\% on both BD and FF, which is expected under this protocol because the unchanged query can still match alternative regularities induced by the shuffled support sets; the relevant diagnostic is the drop relative to the unperturbed AP condition. Under that comparison, FF still drops by about 7 points and BD by about 9 points (Table~\ref{tab:main_results}).

In \emph{Sequence Permutation}, we shuffle the order of drawing actions in the query program while leaving the support examples, query image, and gold label intact. BD accuracy drops by about 5 points relative to AP, but FF drops by about 20 points, showing that ordered stroke composition matters most in the Free-form regime.

The two perturbations affect Basic and Free-form tasks in systematically different ways. 
On Basic (BD) problems, shuffling the assignment of support programs to positive vs.\ negative sets reduces pooled C--G AP accuracy from 68.8\% to 60.1\% (–8.7 points), whereas permuting the action sequence in the query program yields a smaller drop to 64.2\% (–4.6 points), suggesting that coherent support-set structure is more critical than exact stroke order for these library-based shapes. In contrast, on Free-form (FF) problems, where concepts depend more directly on stroke-level geometry, query sequence permutation is far more damaging: accuracy falls from 78.1\% to 58.0\% (–20.1 points), compared to 70.6\% (–7.5 points) under category permutation (more in Appendix~\ref{app:randomization}). 


\section{Discussion}
\label{sec:discussion}

Our results on LOGO reveal a strong dependence of abstract visual reasoning performance on representational format. VLMs operating on raw pixels perform near chance on these Bongard-style tasks, whereas the same underlying reasoning engines, when supplied with structured symbolic input, achieve large gains that, on some categories, approach reported human performance from prior work. This shift, together with the minimal impact of adding a visual anchor and the sharp degradations under structural randomization, highlights a representational bottleneck on Bongard-style tasks: current visual encoders, though successful at object recognition and retrieval, fail to capture the precise geometric and relational structure required for Bongard-style abstraction.

The Componential–Grammatical (C–G) paradigm isolates this representational gap. Symbolic input acts as a grounding scaffold: it preserves object identity, shape structure, and compositional relations that are typically lost or blurred in visual embeddings. Importantly, these representations are interpretable and amenable to structured manipulation, echoing earlier proposals to treat Bongard problems as inference in a visual language of primitives and relations \citep{depeweg2024visual_language, bongard_architecture_2023}, and our grounded C--G experiments show that, within the VLM-capable subset, once such a scaffold is in place additional visual evidence offers at best marginal benefits.

We also observe that input format interacts with task complexity. Natural-language descriptions offer gains on simple categorization tasks (e.g., “contains a triangle”) but degrade on more abstract tasks that require global structure (e.g., symmetry or convexity). Procedural grammars, by contrast, support more reliable reasoning on complex problems. The perturbation results mirror this pattern: BD accuracy is more sensitive to shuffling which programs belong to the positive vs.\ negative sets, whereas FF accuracy collapses primarily when we disrupt the ordered composition of strokes in the query program, indicating that different concept classes rely on different facets of symbolic structure. These format-specific effects suggest that there is no universal “best” representation; alignment between representational form and task structure is critical. Model behavior varies as well. Larger models show strong relative gains from symbolic input but are sensitive to verbosity and format. Smaller, instruction-tuned models can sometimes underperform when exposed to over-structured prompts, highlighting brittleness to task framing (more in Appendix~\ref{app:model_capacity}). Overall, symbolic inputs help most when they preserve structure without overloading the model with linguistic noise, and the randomization controls indicate that it is precisely this relational and compositional structure—rather than superficial token frequencies—that models rely on, especially for stroke-level Free-form tasks.

These results support a modular view of multimodal AI: instead of relying solely on end-to-end systems, it may be more effective to explicitly separate perception from reasoning, using intermediate symbolic representations as a grounding interface. This design pattern is already common in domains such as formal mathematics and program synthesis, and is increasingly reflected in neurosymbolic visual systems: SBSD models Bongard-LOGO concepts via probabilistic concept models over symbolic shape descriptors \citep{sbss_pmoC_2024}, NSA solves ARC tasks by combining neural proposals with search in a domain-specific language \citep{nsa_arc_2025}, agent architectures such as VLAgent plan executable programs from visual input before calling neural modules for execution \citep{xu2025vlagent}, and PRISM explicitly decouples perception and reasoning in VLM pipelines by converting visual information into a textual intermediate representation before downstream inference \citep{qiao2024prism}. Our findings extend this pattern to large language models on Bongard-LOGO, suggesting that general-purpose LLMs function effectively as inference engines once grounded in an appropriate visual language. In this sense, C--G should be viewed as a diagnostic upper bound on what current LLMs can do once grounded in an appropriate visual language, rather than as a proposal for solving perception.

Work on compositional visual reasoning in VLMs points to a complementary axis of progress: unrolling reasoning into multi-step procedures. Progressive alignment methods explicitly model multi-granular interactions between language and image regions, and GFlowVLM shows that CoT style multi-step reasoning improves performance on complex visual tasks \citep{le2025progressive, kang2025gflowvlm}. Neurosymbolic agents similarly decompose queries into symbolic subgoals and verify intermediate results \citep{xu2025vlagent}. Our experiments are orthogonal in that, without changing model internals or prompting them for explicit chains of thought, we demonstrate that simply changing the \emph{representation format} — from pixels to LOGO-style programs or descriptions—already unlocks large gains. Together, these lines of work suggest that multi-step reasoning mechanisms and better grounding interfaces are complementary: CoT procedures operate over whatever representations they are given, and our results indicate that providing a suitable visual language is a powerful lever for improving downstream reasoning.

At the same time, our oracle symbolic interface should not be conflated with natural-image reasoning. In natural-image settings, the intermediate representation itself must be inferred from pixels, typically with noise and partial coverage. Recent representation-first pipelines such as Componential Analysis \citep{vaishnav2025cognitive} and PRISM \citep{qiao2024prism} point toward this direction by first constructing textual or structured world models and then reasoning over them; natural-image Bongard benchmarks such as Bongard-OpenWorld and Bongard-HOI provide testbeds for whether such approximate symbolic interfaces transfer beyond synthetic geometry \citep{wu2024bongardopenworld, jiang2022bongardhoi}. Our results therefore best support a conditional claim: if a model is given a faithful symbolic abstraction, its downstream reasoning can be much stronger than its raw-pixel performance suggests.

Finally, recent “Bongard in Wonderland” evaluations show that even frontier foundation models (e.g., GPT-4o) often fail on elementary Bongard concepts such as spirals \citep{wust2025bongard}. Alongside our symbolic upper bounds and prior visual-language solvers \citep{depeweg2024visual_language, bongard_architecture_2023}, this convergence of evidence supports a common conclusion: robust abstract visual reasoning on Bongard-style benchmarks is unlikely to emerge from monolithic end-to-end VLMs alone, and instead appears to require explicit representational structure, whether via hand-engineered operators, probabilistic visual languages, or program-like grounding interfaces for LLMs.

\section{Limitations}
\label{sec:limitations}

While the C--G approach reveals useful structure–reasoning interactions, it comes with several limitations (expanded in Appendix~\ref{app:limitations}). \textit{First}, Bongard-LOGO is a synthetic benchmark tailored to abstract geometry, which makes it ideal for perturbation studies (grounding, randomization) but leaves open how well our findings about symbolic grounding transfer to natural-image domains such as Bongard-OpenWorld or Bongard-HOI, where the intermediate representation must itself be inferred from cluttered scenes \citep{wu2024bongardopenworld, jiang2022bongardhoi, vaishnav2025cognitive}. 
\textit{Second}, our setup assumes oracle access to ground-truth action programs, effectively removing perception from the loop; in realistic settings such programs would need to be inferred from pixels via inverse graphics or symbolic detectors, and our grounded C--G variants only partially probe this gap, since they still rely on perfect symbolic inputs and add only a single visual anchor. 
\textit{Third}, because we bypass perception entirely, C--G is diagnostic rather than prescriptive: it shows what current models can do \emph{given} suitable structure, but not how to learn such representations, pointing to the need for learned front-ends that induce approximate programs or scene graphs from raw data. 
\textit{Fourth}, our model set, though broad, does not systematically cover architectures with explicit geometric inductive biases, so our observations about capacity and brittleness may not generalize uniformly. 
Finally, even with perfect symbolic input, models still struggle on Human-designed problems involving global structure and multiple abstraction levels, echoing results from ARC and related benchmarks that reasoning over rich concepts remains challenging even for neurosymbolic systems. Inferring symbolic structure from perception, aligning it with downstream tasks, and scaling robust reasoning over such representations remain important directions for future work.

\section{Conclusion}
\label{sec:conclusion}

We have used Bongard-LOGO as a controlled testbed to study how representation shapes abstract visual reasoning in large language models. Comparing end-to-end VLMs on images to the same reasoning engines operating on ground-truth symbolic programs suggests that, on this benchmark, a large portion of the performance gap is attributable to perceptual representations rather than to a lack of reasoning capacity. The Componential--Grammatical (C--G) interface, together with grounded and randomization variants indicates that explicit symbolic structure is the main driver of gains in our experiments, while additional visual anchoring and surface-level token statistics play only a minor role. These findings support a modular view of multimodal AI in which learned or engineered front-ends extract structured visual languages that general-purpose LLMs can then reason over. Extending this approach beyond synthetic settings and developing robust mechanisms for acquiring such representations from pixels remain key directions for future work.

\bibliography{custom}


\onecolumn
\appendix

\setcounter{figure}{0}
\renewcommand{\thefigure}{A.\arabic{figure}}

\setcounter{table}{0}
\renewcommand{\thetable}{A.\arabic{table}}

\setcounter{lstlisting}{0}
\renewcommand{\thelstlisting}{A.\arabic{lstlisting}}

\vspace*{2cm} 
{\centering \Huge \bfseries Appendix \par}
\vspace{1cm} 

\section{Example Bongard-LOGO Instance}
\label{app:example_instance}

To illustrate the different symbolic views used in our experiments, we present a single Bongard-LOGO problem from the Basic (BD) category. The underlying concept combines an unbalanced trapezoid\/right triangle configuration with an uneven band of four arcs.

\subsection{Concept and Natural-Language Description}

This problem is labeled with the high-level Basic concept:

\begin{itemize}
    \item \textbf{Concept UI:} ``unbalanced trapezoid right\_triangle AND uneven band four arcs''
\end{itemize}

\subsection{Visual Layout}

\begin{figure}[htb]
    \centering
    \setlength{\fboxsep}{0pt} 
    \setlength{\fboxrule}{1pt} 
    
    \begin{subfigure}[b]{0.45\linewidth}
        \centering
        \fbox{\includegraphics[width=\linewidth]{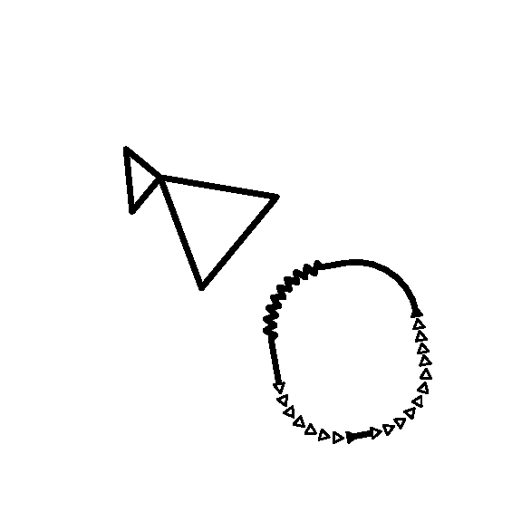}}
        \caption{Negative example}
        \label{fig:neg_example}
    \end{subfigure}
    \hfill
    \begin{subfigure}[b]{0.45\linewidth}
        \centering
        \fbox{\includegraphics[width=\linewidth]{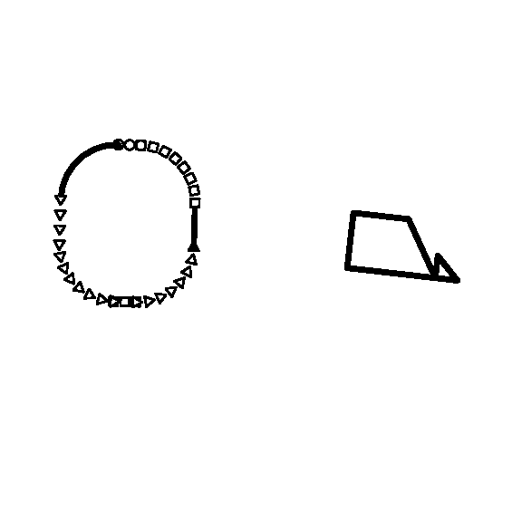}}
        \caption{Positive example}
        \label{fig:pos_example}
    \end{subfigure}
    
    \caption{
    Example Basic (BD) Bongard-LOGO problem. Both the images conforming to the ``unbalanced trapezoid right\_triangle AND uneven band four arcs'' concept.}
    \label{fig:example_bd_images}
\end{figure}

Figure~\ref{fig:example_bd_images} shows the one positive and one negative example image.

\subsection{Procedural Action Programs}
\label{app:eg_action_program}

Table~\ref{tab:example_bd_programs} shows the LOGO-style action programs for one negative and one positive example. Each token encodes both a primitive type and quantized geometric parameters (e.g., length, radius, angle).

\begin{table}[htb]
    \centering
    \caption{Action programs for one negative and one positive example in the BD instance.}
    \label{tab:example_bd_programs}
    \begin{tabular}{p{0.4\linewidth}p{0.4\linewidth}}
    \toprule
    \textbf{Negative Action Program} & \textbf{Positive Action Program} \\
    \midrule
    \texttt{line\_normal\_0.300-0.500}, & \texttt{line\_normal\_1.000-0.500}, \\
    \texttt{line\_normal\_0.424-0.875}, & \texttt{line\_normal\_0.283-0.875}, \\
    \texttt{line\_normal\_0.300-0.875}, & \texttt{line\_normal\_0.200-0.875}, \\
    \texttt{line\_normal\_0.800-0.167}, & \texttt{line\_normal\_0.583-0.086}, \\
    \texttt{line\_normal\_0.800-0.833}, & \texttt{line\_normal\_0.500-0.664}, \\
    \texttt{line\_normal\_0.800-0.833}, & \texttt{line\_normal\_0.500-0.750}, \\
    \texttt{line\_normal\_0.200-0.500}, & \texttt{line\_square\_0.200-0.500}, \\
    \texttt{arc\_zigzag\_0.500\_0.625-0.500}, & \texttt{arc\_triangle\_0.500\_0.625-0.500}, \\
    \texttt{line\_normal\_0.400-0.500}, & \texttt{line\_normal\_0.400-0.500}, \\
    \texttt{arc\_triangle\_0.500\_0.625-0.500}, & \texttt{arc\_square\_0.500\_0.625-0.500}, \\
    \texttt{line\_normal\_0.200-0.500}, & \texttt{line\_circle\_0.200-0.500}, \\
    \texttt{arc\_triangle\_0.500\_0.625-0.500}, & \texttt{arc\_normal\_0.500\_0.625-0.500}, \\
    \texttt{line\_triangle\_0.400-0.500}, & \texttt{line\_triangle\_0.400-0.500}, \\
    \texttt{arc\_normal\_0.500\_0.625-0.500} & \texttt{arc\_triangle\_0.500\_0.625-0.500} \\
    \bottomrule
    \end{tabular}
\end{table}

\subsection{Natural-Language Action Descriptions}
\label{app:eg_action_description}

For the C--G (Action Description) condition, each program is rendered as a stepwise English procedure. Below we show the corresponding descriptions for the same negative and positive examples.

\paragraph{Negative example.}

\begin{quote}
To draw figure 1, follow these steps:

Step 1: draw a normal line of 0.300 units. Then, continue straight without turning.

Step 2: draw a normal line of 0.424 units. After that, turn left by 135.0 degrees.

Step 3: draw a normal line of 0.300 units. After that, turn left by 135.0 degrees.

Step 4: draw a normal line of 0.800 units. After that, turn right by 119.9 degrees.

Step 5: draw a normal line of 0.800 units. After that, turn left by 119.9 degrees.

Step 6: draw a normal line of 0.800 units. After that, turn left by 119.9 degrees.

Step 7: draw a normal line of 0.200 units. Then, continue straight without turning.

Step 8: draw a zigzag arc with a radius of 0.500 and sweeping 90.0 degrees. Then, continue straight without turning.

Step 9: draw a normal line of 0.400 units. Then, continue straight without turning.

Step 10: draw a triangle arc with a radius of 0.500 and sweeping 90.0 degrees. Then, continue straight without turning.

Step 11: draw a normal line of 0.200 units. Then, continue straight without turning.

Step 12: draw a triangle arc with a radius of 0.500 and sweeping 90.0 degrees. Then, continue straight without turning.

Step 13: draw a triangle line of 0.400 units. Then, continue straight without turning.

Step 14: draw a normal arc with a radius of 0.500 and sweeping 90.0 degrees. Then, continue straight without turning.

The figure is now complete.
\end{quote}

\paragraph{Positive example.}

\begin{quote}
To draw figure 1, follow these steps:

Step 1: draw a normal line of 1.000 units. Then, continue straight without turning.

Step 2: draw a normal line of 0.283 units. After that, turn left by 135.0 degrees.

Step 3: draw a normal line of 0.200 units. After that, turn left by 135.0 degrees.

Step 4: draw a normal line of 0.583 units. After that, turn right by 149.0 degrees.

Step 5: draw a normal line of 0.500 units. After that, turn left by 59.0 degrees.

Step 6: draw a normal line of 0.500 units. After that, turn left by 90.0 degrees.

Step 7: draw a square line of 0.200 units. Then, continue straight without turning.

Step 8: draw a triangle arc with a radius of 0.500 and sweeping 90.0 degrees. Then, continue straight without turning.

Step 9: draw a normal line of 0.400 units. Then, continue straight without turning.

Step 10: draw a square arc with a radius of 0.500 and sweeping 90.0 degrees. Then, continue straight without turning.

Step 11: draw a circle line of 0.200 units. Then, continue straight without turning.

Step 12: draw a normal arc with a radius of 0.500 and sweeping 90.0 degrees. Then, continue straight without turning.

Step 13: draw a triangle line of 0.400 units. Then, continue straight without turning.

Step 14: draw a triangle arc with a radius of 0.500 and sweeping 90.0 degrees. Then, continue straight without turning.

The figure is now complete.
\end{quote}

\subsection{Concept-Conditioned Prompt Fragment}
\label{app:eg_concept}

When evaluating the Concept-Conditioned C--G setting, the same instance is annotated with a high-level rule:

\begin{quote}
\textbf{Concept (system prompt fragment):}  
``Here is the overall concept behind the positive samples:  
\emph{unbalanced trapezoid right\_triangle AND uneven band four arcs.}''
\end{quote}

This example illustrates the three kinds of input used in our experiments for a single Bongard-LOGO problem: raw images (Figure~\ref{fig:example_bd_images}), procedural action programs (Table~\ref{tab:example_bd_programs}), and natural-language action descriptions, optionally augmented with an explicit concept label.


\section{Model Details and Version Information}
\label{app:models}

\subsection{Proprietary Models}

\begin{table}[htbp]
\centering
\small
\caption{Proprietary Models Evaluated}
\label{tab:proprietary_models}
\begin{tabular}{lllll}
\toprule
\textbf{Model} & \textbf{Developer} & \textbf{Release} & \textbf{Context Window} & \textbf{Multi-Image} \\
\midrule
Gemini-2.5-flash-preview & Google & May 2025 & 1M tokens & Yes \\
Gemini-3-flash-preview & Google & Dec. 2025 & 1M tokens & Yes \\
\bottomrule
\end{tabular}
\end{table}

\textbf{Access:} Google AI Studio APIs. Requires authentication and API key.

\subsection{Open-Source Models (via Ollama)}

\textbf{Infrastructure:} All open-source models executed locally using Ollama framework on HPC cluster with heterogeneous NVIDIA GPUs (A100 40GB/80GB VRAM).


\section{Reproducibility and Implementation Details}
\label{app:reproducibility}

\subsection{Hardware Requirements}

\textbf{Minimum Configuration:}
\begin{itemize}
    \item GPU: NVIDIA RTX 3090 (24GB VRAM) or equivalent
    \item RAM: 64GB system memory
    \item Storage: 500GB for models and datasets
\end{itemize}

\subsection{Software Stack}

\begin{itemize}
    \item Ollama v0.9.6 or later (model serving framework)
    \item Python 3.10+
    \item PyYAML for configuration parsing
    \item Requests library for HTTP communication
    \item Google API Python client for Gemini API access
\end{itemize}

\subsection{Dataset Access}

\textbf{Bongard-LOGO Dataset:}
Available from original authors at: \url{https://github.com/NVlabs/Bongard-LOGO}

Dataset includes:
\begin{itemize}
    \item 12,000 problem instances (3,600 Free-form, 4,000 Basic, 4,400 Abstract)
    \item Ground-truth labels and rule annotations
    \item Action program specifications for all images
    \item Human-designed shape library with 627 shapes and action programs
\end{itemize}

\subsection{Code Availability}

Complete experimental code, including:
\begin{itemize}
    \item Data loading and preprocessing utilities
    \item Prompt generation functions (all variants in Appendix~\ref{app:prompts})
    \item Model interaction wrappers (Ollama + Gemini API)
    \item Evaluation and result aggregation scripts
    \item Statistical analysis and visualization code
\end{itemize}

Code will be released upon paper acceptance at a public GitHub repository.

\paragraph{API usage for grounded calls.}
Grounded C--G calls differ from symbolic-only C--G only in that the query image is passed alongside the textual prompt. For Gemini models, we pass the prompt and a PIL image object; for Ollama-served VLMs, we add the image as a base64-encoded field in the system message. This setup ensures that \emph{only} the query receives both symbolic and visual input, while positives and negatives remain purely symbolic, making Grounded C--G a minimal-anchoring ablation on top of the Componential--Grammatical interface.

\subsection{Computational Costs}

This work required substantial computational resources:
\begin{itemize}
    \item 2000+ GPU-hours on HPC infrastructure
\end{itemize}


\section{Complete Prompts for All Experimental Conditions}
\label{app:prompts}

This appendix provides the complete, unedited prompts used for each experimental condition. These prompts are essential for reproducibility and can be adapted for similar reasoning tasks.

\subsection{Baseline: End-to-End VLM Visual Reasoning}
\label{app:prompt_vlm}

\begin{lstlisting}[language=Python, caption={VLM Visual Reasoning Prompt}, label=lst:prompt_vlm]
SYSTEM_PROMPT = """
You are an expert visual analyst, working on Bongard Logo problems containing synthetic images with abstract shapes. You are presented with {n + m + 1} images:
    - The first {n} images are labeled 'cat_2' (positive examples).
    - The next {m} images are labeled 'cat_1' (negative examples).
    - The last image is the 'Test Image'.

    Your task is to analyze the visual features of all images and then classify the 'Test Image' based on the features it shares with 'cat_1' or 'cat_2'. 

    **Your Task:**

    1.  **Analyze Positive and Negative Images:**
        *   Carefully compare the positive (cat_2) and negative (cat_1) images.
        *   **Describe the visual characteristics of *each* category (cat_1 and cat_2) in detail.**  Focus on shapes, lines, orientations, spatial relationships, and quantities.  Break down complex shapes into simpler components. Describe shapes as if explaining them to someone who can't see them.
        *   The goal is to implicitly understand the pattern distinguishing cat_2 from cat_1, even if you don't explicitly state it as a single rule.

    2.  **Analyze the Test Image:**
        *   Describe the test image's visual features in detail, using the same level of detail as in step 1.

    3.  **Classify the Test Image:**
        *   Based on your analysis of the positive, negative, and test images, determine whether the test image belongs to cat_2 (positive) or cat_1 (negative).
        *   **Provide a clear justification for your classification.** Explain *why* the test image's features align more closely with the positive examples or the negative examples, referencing your detailed descriptions from steps 1 and 2.

    **Output Format:**

    *   **Analysis of Positive and Negative Examples:** (Detailed descriptions of cat_1 and cat_2 characteristics)
    *   **Test Image Analysis:** (Detailed description of the test image's features)
    *   **Test Image Classification:** (Either "cat_2 (Positive)" or "cat_1 (Negative)", followed by your justification)

    **Begin Analysis:**
    **Output Format (JSON):**
    }
        "Test Image": "(Detailed description of the test image's features)",
        "Rule": "(List of features identified as consistently present in cat_2 category)",
        "Analysis": "(Detailed descriptions of cat_1 and cat_2 characteristics)",
        "Conclusion": "(cat_1 or cat_2)"
    }

    Respond in the JSON format provided, and avoid unnecessary explanations or text outside of the JSON structure.

    **Begin Analysis of All Images:**
"""
\end{lstlisting}

\subsection{Componential-Grammatical (C-G): Action Programs}
\label{app:prompt_cg}

\begin{lstlisting}[language=Python, caption={C-G Action Program Prompt}, label=lst:prompt_cg]
SYSTEM_PROMPT = """
**Objective:** Solve a Bongard-style visual reasoning puzzle by identifying the rule that distinguishes a "positive" set of programs from a "negative" set.

        **Input Context:**
        - You will receive human-readable, step-by-step instructions that describe how to construct a set of abstract shapes. These are called "action descriptions".
        - You will NOT be shown any images. Your task is to reason directly from these textual descriptions.

        **Your Core Task:**
        1.  Analyze the provided "action descriptions" for the positive, negative, and test sets.
        2.  Infer the final geometric properties of the shapes from these descriptions.
        3.  Identify the abstract rule that defines the positive set.
        4.  **Crucially, your reasoning must focus on the final geometric properties of the shapes, not the specific steps used to construct them.**

        Here is the overall concept behind the positive samples: {concept}

        **Your Task and Required Output:**
        Respond **only** with a single JSON object. Do not include any text or explanations outside of the JSON structure.

        The JSON object must have the following keys:
        }
            "Analysis": "A brief analysis of the distinguishing characteristics you observed in the 'pos' samples that make them distinct from 'neg' samples.",
            "Rule": "A clear and concise rule that defines the positive samples.",
            "Test Image": "A summary of the test sample's properties, as inferred from its description, and how it fits or doesn't fit the derived rule.",
            "Conclusion": "Your final categorization of the test image. The value must be either 'pos' or 'neg'."
        }
"""
\end{lstlisting}

\subsection{Concept-Conditioned C--G}
\label{app:prompt_concept}
We add a single declarative sentence specifying the target concept, keeping the rest of the prompt identical across experiments to isolate the effect of explicit concept conditioning. Concept-Conditioned C--G uses the same prompt as in Listing~\ref{lst:prompt_cg}, with one additional line inserted before the “Your Task and Required Output” block:

\begin{description}
    \item[Here is the overall concept behind the positive samples: \{\textit{concept}\}]
\end{description}

\subsection{Natural Language Procedures (Ablation)}
\label{app:prompt_nl}

\begin{lstlisting}[language=Python, caption={Natural Language Action Description Prompt}, label=lst:prompt_ad]
SYSTEM_PROMPT = """
**Objective:** Solve a Bongard-style visual reasoning puzzle by identifying the rule that distinguishes a "positive" set of programs from a "negative" set.

        **Input Context:**
        - You will receive human-readable, step-by-step instructions that describe how to construct a set of abstract shapes. These are called "action descriptions".
        - You will NOT be shown any images. Your task is to reason directly from these textual descriptions.

        **Your Core Task:**
        1.  Analyze the provided "action descriptions" for the positive, negative, and test sets.
        2.  Infer the final geometric properties of the shapes from these descriptions.
        3.  Identify the abstract rule that defines the positive set.
        4.  **Crucially, your reasoning must focus on the final geometric properties of the shapes, not the specific steps used to construct them.**

        **Your Task and Required Output:**
        Respond **only** with a single JSON object. Do not include any text or explanations outside of the JSON structure.

        The JSON object must have the following keys:
        {
            "Analysis": "A brief analysis of the distinguishing characteristics you observed in the 'pos' samples that make them distinct from 'neg' samples.",
            "Rule": "A clear and concise rule that defines the positive samples.",
            "Test Image": "A summary of the test sample's properties, as inferred from its description, and how it fits or doesn't fit the derived rule.",
            "Conclusion": "Your final categorization of the test image. The value must be either 'pos' or 'neg'."
        }
"""
\end{lstlisting}

\subsection{Minimal-Context Baseline (Ablation)}
\label{app:prompt_minimal}

\begin{lstlisting}[language=Python, caption={Minimal-Context Prompt}, label=lst:prompt_minimal]
SYSTEM_PROMPT = """
**Your Task and Required Output:**
        Analyze the provided samples and respond *only* with a single JSON object. Do not include any additional text, explanations, or markdown formatting outside of the JSON structure.

        The JSON object must have the following keys:
        }
            "Analysis": "A brief analysis of the distinguishing characteristics you observed in the 'pos' samples that make them distinct from 'neg' samples.",
            "Rule": "A clear and concise rule that defines the positive samples.",
            "Test Image": "A summary of the test image's properties as they relate to the derived rule.",
            "Conclusion": "Your final categorization of the test image. The value must be either 'pos' or 'neg'."
        }
"""
\end{lstlisting}

\subsection{Grounded Componential--Grammatical (Ablation)}
\label{app:prompt_grounded}

In the Grounded C--G setting, models receive the same symbolic inputs as in the main C--G conditions, but the \emph{query} also includes its rendered image. The goal is to test whether a single visual anchor, paired with otherwise symbolic input, provides additional benefits once an explicit visual language is available.

Concretely:
\begin{itemize}
    \item For Gemini models, we call the official API with a text prompt and a PNG image object as a second content element.
    \item For open VLMs served via Ollama (e.g., Gemma3, LLaVA-LLaMA3, Qwen2.5-VL), we send a system message containing the text prompt and attach a single image encoded as base64.
\end{itemize}
In both cases, only the query image is passed; positive and negative examples are provided in symbolic form (action programs), and the model must output a JSON object with an analysis, an induced rule, a description of the test image, and a binary conclusion (\texttt{pos}/\texttt{neg}).

\paragraph{Grounded base (visual-only baseline).}
For the purely visual “base” context, the system prompt is:

\begin{lstlisting}[language=Python, caption={Minimal-Context Prompt}, label=lst:prompt_ground_base]
SYSTEM_PROMPT = """
    **Your Task and Required Output:**
    Analyze the provided samples and respond *only* with a single JSON object. Do not include any additional text, explanations, or markdown formatting outside of the JSON structure.

    The JSON object must have the following keys:
    }
        "Analysis": "A brief analysis of the distinguishing characteristics you observed in the 'pos' samples that make them distinct from 'neg' samples.",
        "Rule": "A clear and concise rule that defines the positive samples.",
        "Test Image": "A summary of the test image's attributes, referencing the provided png image.",
        "Conclusion": "Your final categorization of the test image. The value must be either 'pos' or 'neg'."
    }
    """
\end{lstlisting}

\paragraph{Grounded C--G with Action Descriptions.}
For the grounded Action Description condition, the model sees symbolic descriptions for all examples and a \emph{paired} description+image for the query:

\begin{lstlisting}[language=Python, caption={Minimal-Context Prompt}, label=lst:prompt_ground_ad]
SYSTEM_PROMPT = """
    **Persona:** You are an expert visual reasoning system specializing in analyzing abstract patterns.

    **Objective:** Your primary goal is to solve a Bongard-style visual reasoning puzzle. You will be given abstract descriptions (action programs) used to generate synthetic images and a reference image. You must identify the underlying rule that distinguishes a "positive" set from a "negative" set.

    **Crucially, for the test sample, you will be given both its action program and the final rendered image.** Use this pair to understand the connection between the symbolic action programs and their visual generation. Apply this understanding to decipher the rule governing the positive and negative sets, for which you only have action programs. These action programs are the steps to draw the abstract shape on the canvas.

    **Your Task and Required Output:**
    Analyze the provided samples and respond *only* with a single JSON object. Do not include any additional text, explanations, or markdown formatting outside of the JSON structure.

    The JSON object must have the following keys:
    }
        "Analysis": "A brief analysis of the distinguishing characteristics you observed in the 'pos' samples that make them distinct from 'neg' samples.",
        "Rule": "A clear and concise distinguishing rule that defines the positive samples.",
        "Test Image": "A summary of the test image's attributes, referencing the provided png image.",
        "Conclusion": "Your final categorization of the test image. The value must be either 'pos' or 'neg'."
    }
    """
\end{lstlisting}

\paragraph{Grounded C--G with Action Programs.}
For the grounded Action Program condition, the model receives LOGO-style programs for all examples and a program+image pair for the query, with an extended explanation of the program format:

\begin{lstlisting}[language=Python, caption={Minimal-Context Prompt}, label=lst:prompt_ground_ap]
SYSTEM_PROMPT = """
    **Persona:** You are an expert visual reasoning system specializing in analyzing abstract patterns.

    **Objective:** Your primary goal is to solve a Bongard-style visual reasoning puzzle. You will be given abstract descriptions (action programs) used to generate synthetic images. You must identify the underlying rule that distinguishes a "positive" set from a "negative" set.

    **Input Format and Generative Context:**
    You will receive 13 sets of abstract figure descriptions enclosed in [...]. These descriptions are the **source code** (called "action programs") that programmatically generate the images. Understanding the generation process is key.

    *   **The Generation Pipeline:**
        1.  **Stage 1: Base Shape Definition:** Shapes are first defined with basic geometry using degrees for angles.
            *   `set of base actions`: A list of primitives like `line_LENGTH` or `arc_RADIUS_ANGLE`.
            *   `turn angles`: A sequence of turns like `L90.0--R45.0...`, specified in degrees.
        2.  **Stage 2: Object Creation (`BasicAction`):** The system reads the base geometry and creates programmable `BasicAction` objects. At this stage, a visual stroke style (`line_type` like "zigzag" or "triangle") is added.
        3.  **Stage 3: Final Serialization (The Input You See):** The `BasicAction` objects are serialized into the final action program strings. All values are **normalized** into a [0, 1] range.
            *   **Line String:** `line_TYPE_LENGTH-TURNANGLE`
            *   **Arc String:** `arc_TYPE_ARCANGLE_ARCRADIUS-TURNANGLE`

    *   **Program Structure Hierarchy:**
        The final action program is a nested list: `BongardProblem` -> `BongardImage` -> `OneStrokeShape` -> `BasicAction`.

    *   **Data Sets You Will Analyze:**
        1.  **Positive Samples (`pos`):** Action programs that all conform to a common abstract rule.
        2.  **Negative Samples (`neg`):** Action programs that do not conform to this rule.
        3.  **Test Sample:** An action program to be categorized, accompanied by its rendered image to provide visual context.

    *   **Crucial Interpretive Note:** While the action programs detail *how* a figure is constructed, your task is to reason about the final, abstract geometric properties of the resulting figure. The construction method is just one of several possible ways to draw the same shape.

    **Your Task and Required Output:**
    Analyze the provided samples and respond *only* with a single JSON object. Do not include any additional text, explanations, or markdown formatting outside of the JSON structure.

    The JSON object must have the following keys:
    }
        "Analysis": "A brief analysis of the distinguishing characteristics you observed in the 'pos' samples that make them distinct from 'neg' samples.",
        "Rule": "A clear and concise distinguishing rule that defines the positive samples.",
        "Test Image": "A summary of the test image's attributes, referencing the provided png image.",
        "Conclusion": "Your final categorization of the test image. The value must be either 'pos' or 'neg'."
    }
    """
\end{lstlisting}

\subsection{User Prompt}
\label{app:user_prompt}
We used the same user prompt for all our experiments. 

\begin{lstlisting}[language=Python, caption={Natural Language Procedure Prompt}, label=lst:user_prompt]
USER_PROMPT = """
POSITIVE SET (6 descriptions):
[Description 1]
[Description 2]
[Description 3]
[Description 4]
[Description 5]
[Description 6]

NEGATIVE SET (6 descriptions):
[Description 7]
[Description 8]
[Description 9]
[Description 10]
[Description 11]
[Description 12]

QUERY (1 description):
[Description 13]

Classify the query as 'positive' or 'negative'. Respond with JSON 
only.
"""
\end{lstlisting}


\section{Extended Results Tables}
\label{app:results_tables}

\subsection{Complete Results by Model and Condition}

Table~\ref{tab:bongard_results} reports full C--G performance on the Bongard-LOGO benchmark across all 12 models and experimental conditions.
For each model, we include Action Description, Action Program, and minimal-context Base (Ablation) conditions, broken down by Basic Shapes (BD), Free-form (FF), Human-designed Combined (HD Comb), and Human-designed Novel (HD Novel).

\begin{table}[htbp]
\centering
\caption{C-G paradigm performance on Bongard-LOGO.
Results show accuracy (\%) for Action Desc. and Action Prog. across problem categories with concept ($C$) and without concept ($NC$) guidance.
BD = Basic Shapes, FF = Free-form, HD = Human-designed. Action Desc. = Action description and Action Prog. = Action program.}
\label{tab:bongard_results}
\begin{tabular}{l l S[table-format=2.1] S[table-format=2.1] S[table-format=2.1] S[table-format=2.1] S[table-format=2.1] S[table-format=2.1] S[table-format=2.1] S[table-format=2.1]}
\toprule
\multirow{2}{*}{Model} & \multirow{2}{*}{Experiment} & \multicolumn{2}{c}{BD} & \multicolumn{2}{c}{FF} & \multicolumn{2}{c}{HD Comb} & \multicolumn{2}{c}{HD Novel} \\
\cmidrule(lr){3-4} \cmidrule(lr){5-6} \cmidrule(lr){7-8} \cmidrule(lr){9-10}
 & & {$C$} & {$NC$} & {$C$} & {$NC$} & {$C$} & {$NC$} & {$C$} & {$NC$} \\
\midrule
\multicolumn{9}{l}{\textit{DeepSeek-R1}} \\
\multirow{2}{*}{8b} & Action Desc. & 61.8 & 65.6 & 61.6 & 63 & 50.6 & 56.2 & 52.2 & 54.6 \\
 & Action Prog. & 56 & 57.52 & 56 & 56.2 & 53.4 & 54.91 & 55.2 & 53.8 \\
 & Baseline & \- & 53.0 & \- & 51.4 & \- & 48.8 & \- & 53.0 \\
\addlinespace\cline{2-10}\addlinespace
\multirow{2}{*}{14b} & Action Desc. & 70.6 & 70.4 & 75.4 & 74 & 58.6 & 57 & 58.6 & 57.6 \\
 & Action Prog. & 69.6 & 68.8 & 75.6 & 77.8 & 56.8 & 57.8 & 59 & 57.6 \\
 & Baseline & \- & 71.8 & \- & 74.6 & \- & 59.2 & \- & 58.6 \\
\addlinespace\cline{2-10}\addlinespace
\multirow{2}{*}{32b} & Action Desc. & 69.2 & 64.6 & 78.2 & 82.8 & 55.6 & 59.6 & 59.6 & 58.6 \\
 & Action Prog. & 67.4 & 65.2 & 81.2 & 78.6 & 59.4 & 57.8 & 59.2 & 64.2 \\
 & Baseline & \- & 66.2 & \- & 74.6 & \- & 55.8 & \- & 58.2 \\
\midrule
\multicolumn{9}{l}{\textit{Gemma3}} \\
\multirow{2}{*}{12b} & Action Desc. & 65 & 65.06 & 72.8 & 69.2 & 58.6 & 56.6 & 61 & 65.8 \\
 & Action Prog. & 62.8 & 64 & 71.4 & 70 & 63 & 65.4 & 62.4 & 63.6 \\
 & Baseline & \- & 66.6 & \- & 71.2 & \- & 60.8 & \- & 62.2 \\
\addlinespace\cline{2-10}\addlinespace
\multirow{2}{*}{27b} & Action Desc. & 70.2 & 65.6 & 79.4 & 81 & 64.2 & 57.4 & 70 & 62.6 \\
 & Action Prog. & 69.4 & 67 & 87.4 & 83.2 & 64.4 & 66 & 64.4 & 66.6 \\
 & Baseline & \- & 67.6 & \- & 82.8 & \- & 63.0 & \- & 66.8 \\
\midrule
\multirow{2}{*}{\textit{Magistral}:24b} & Action Desc. & 73.8 & 75.8 & 85.8 & 84.8 & 60.6 & 60.4 & 66 & 63.2 \\
 & Action Prog. & 72.8 & 70.6 & 81.6 & 80.4 & 57.8 & 60 & 67.2 & 58.2 \\
 & Baseline & \- & 72.6 & \- & 82.2 & \- & 59.4 & \- & 64.2 \\
\midrule
\multirow{2}{*}{\textit{Phi4}:14b} & Action Desc. & 68.8 & 75.2 & 91.8 & 90.2 & 58.8 & 55 & 59.6 & 58 \\
 & Action Prog. & 71.6 & 70.8 & 92 & 89.8 & 60.2 & 62.2 & 64.8 & 62.4 \\
 & Baseline & \- & 70.2 & \- & 97.0 & \- & 61.2 & \- & 61.4 \\
\midrule
\multirow{2}{*}{\textit{Phi4-Reasoning}:14b} & Action Desc. & 73.4 & 79.4 & 97.4 & 87.8 & 57.2 & 51.8 & 60.4 & 58.8 \\
 & Action Prog. & 75.6 & 72.6 & 94.2 & 96.2 & 60 & 59.4 & 62.8 & 63.2 \\
 & Baseline & \- & 70.2 & \- & 97.0 & \- & 55.6 & \- & 59.0 \\
\midrule
\multirow{2}{*}{\textit{Qwen2.5}:32b} & Action Desc. & 71.8 & 73.8 & 75 & 80.8 & 60.8 & 60.2 & 64.2 & 61.4 \\
 & Action Prog. & 74.4 & 73.6 & 78.2 & 76.2 & 63.6 & 62.2 & 65 & 63 \\
 & Baseline & \- & 71.4 & \- & 75.8 & \- & 63.4 & \- & 60.8 \\
\midrule
\multirow{2}{*}{\textit{Qwen2.5vl}:32b} & Action Desc. & 58.4 & 68.6 & 70.8 & 72.8 & 59 & 59.6 & 58.6 & 59.4 \\
 & Action Prog. & 67.2 & 66.8 & 73 & 72.4 & 62.6 & 60.2 & 61 & 61.2 \\
 & Baseline & \- & 60.2 & \- & 65.6 & \- & 60.8 & \- & 56.2 \\
\midrule
\multicolumn{9}{l}{\textit{Qwen 3}} \\
\multirow{2}{*}{30b} & Action Desc. & 74.4 & 75.2 & 84 & 78.2 & 60.6 & 56.8 & 62.6 & 63.2 \\
 & Action Prog. & 75.6 & 72.6 & 79.2 & 77.4 & 59.4 & 59.4 & 60.8 & 61.4 \\
 & Baseline & \- & 65.4 & \- & 74.0 & \- & 58.6 & \- & 61.4 \\
\addlinespace\cline{2-10}\addlinespace
\multirow{2}{*}{32b} & Action Desc. & 78.4 & 84.6 & 81.6 & 86.8 & 61.6 & 58.8 & 60.6 & 64.6 \\
 & Action Prog. & 77.6 & 76.4 & 82.2 & 78.8 & 61.8 & 61 & 62.4 & 61.4 \\
 & Baseline & \- & 77.4 & \- & 81.8 & \- & 59.4 & \- & 63.4 \\
\bottomrule
\end{tabular}
\end{table}

\subsection{Effect of Concept Conditioning by Model}
\label{app:concept_results}

Table~\ref{tab:bongard_results} details the impact of concept conditioning for each model, representation, and category.
These numbers underlie the aggregated trends: concept conditioning has small, model- and category-dependent effects, with Action Programs often showing slightly more stable behavior than Action Descriptions in the Human-designed regime.




Concept conditioning yields differential benefits: Action Program representations benefit more (+3.2\% avg) than Action Description (+1.3\%). Abstract Shape problems show largest gains (+5.9\% for AP), Free-form problems minimal gain (+0.9\%).

\subsection{Effect of Symbolic Representation Format}
\label{app:representation}

Natural-language descriptions provide strong inductive bias for the easier FF and BD regimes but fail to generalize to Human-designed abstractions. In contrast, action programs yield smaller but consistent gains, motivating their use as the primary concept representation. Overall, format choice has a modest impact relative to the much larger gap between symbolic and visual-only input.

\subsection{Effect of Model Capacity}
\label{app:model_capacity}
We observe that the effect of symbolic input and concept conditioning varies across models (Table~\ref{tab:model_size_effect}). While some smaller and mid-sized models show clear gains over their visual baselines, others see little change or even slight decreases, and the larger models in our analysis tend to exhibit modest but more consistently positive improvements. This pattern is compatible with the view that structured input can act as an inductive bias that reduces representational burden, but its benefits are not uniform across capacity tiers.

\begin{table}[htb]
\centering
\small
\caption{Mean accuracy improvement (\%) over baseline across all regimes,
stratified by model.
The effect of symbolic input and concept conditioning is heterogeneous: larger models show slightly more consistent positive averages, while smaller and instruction-sensitive models exhibit mixed gains.}
\label{tab:model_size_effect}
\begin{tabular}{lcccc}
\toprule
\textbf{Model} & 
\textbf{AD} & 
\textbf{AP} & 
\textbf{AD + C} & 
\textbf{AP + C} \\
\midrule
DeepSeek-R1:14b & -0.7 & -0.3 & +0.0 & -0.5 \\
DeepSeek-R1:32b & +3.6 & +3.1 & +1.7 & +2.8 \\
DeepSeek-R1:8b & +8.0 & +3.5 & +4.8 & +3.4 \\
Gemini-3-flash-preview & +0.3 & +1.3 & -- & -- \\
Gemma3:12b & -0.4 & +0.7 & -0.6 & +0.0 \\
Gemma3:27b & -2.9 & +0.8 & -0.4 & +1.1 \\
Magistral & +0.0 & -2.5 & +1.2 & -0.5 \\
Phi4 & -4.1 & -1.8 & -3.5 & -1.1 \\
Phi4-Reasoning & -1.7 & +2.5 & +1.1 & +2.2 \\
Qwen2.5:32b & +2.1 & +1.6 & +0.5 & +2.9 \\
Qwen2.5vl:32b & +4.5 & +4.4 & +0.4 & +4.6 \\
Qwen3:30b & -0.1 & +2.3 & +2.0 & +0.3 \\
Qwen3:32b & +2.4 & -1.3 & +2.2 & +2.7 \\
\bottomrule
\end{tabular}
\end{table}

To analyze how symbolic input interacts with model capacity, we partition models into \emph{smaller} and \emph{larger} capacity tiers.
This grouping is based on a combination of parameter scale,
reasoning-oriented pretraining, and baseline performance on Bongard-LOGO.
The grouping is fixed across all experiments and comparisons.
\begin{table}[ht]
\centering
\small
\caption{
Model grouping used for capacity-based analysis.
Models are grouped by approximate parameter scale and reasoning capacity.
}
\label{tab:model_groups}
\begin{tabular}{ll}
\toprule
\textbf{Capacity Tier} & \textbf{Models} \\
\midrule
Smaller &
Phi-4-Reasoning, Phi-4, Magistral, Gemma3:12B, \\
& DeepSeek-R1:7B, DeepSeek-R1:14B \\
\addlinespace
Larger &
Qwen3:32B, Qwen3:30B, Gemma3:27B, DeepSeek-R1:32B, \\& Qwen2.5vl:32B, Qwen2.5:32B \\
\bottomrule
\end{tabular}
\end{table}

To further analyze the interaction between model capacity and symbolic input, we group models into smaller and larger tiers and average accuracy improvements within each group. This grouping is based on a combination of parameter scale as shown in Table~\ref{tab:model_groups}. Accuracy is presented in Table~\ref{tab:size_averaged_gains}. Larger models in this dataset show slightly higher mean gains from symbolic input and concept conditioning than the smaller group, although variance across individual models remains substantial. These exploratory results suggest that the benefits of symbolic input depend on both capacity and pretraining, rather than uniformly favoring lower-capacity models.

\begin{table}[ht]
\centering
\small
\caption{
Mean accuracy improvement (\%) over baseline, averaged across model size groups.
In this dataset, larger models show slightly higher mean gains from symbolic input and concept conditioning than the smaller group.
}
\label{tab:size_averaged_gains}
\begin{tabular}{lcccc}
\toprule
\textbf{Model Group} &
\textbf{AD} &
\textbf{AP} &
\textbf{AD + Concept} &
\textbf{AP + Concept} \\
\midrule
Smaller Models &
$\mathbf{+0.2 \pm 3.7}$ &
$\mathbf{+0.4 \pm 2.2}$ &
$\mathbf{+0.5 \pm 2.5}$ &
$\mathbf{+0.6 \pm 1.6}$ \\
Larger Models &
$+1.6 \pm 2.5$ &
$+1.8 \pm 1.8$ &
$+1.1 \pm 0.9$ &
$+2.4 \pm 1.4$ \\
\bottomrule
\end{tabular}
\end{table}

\subsection{Effect of Grounded C--G with Query Image Anchors}
\label{app:grounded_cg}

The Grounded C--G condition augments symbolic input with a single visual anchor: the rendered query image. This condition is only applicable to models that support image inputs. We therefore restrict this analysis to VLM-capable models: Gemma3 12B, Gemma3 27B, LLaVA-Llama3, and Qwen2.5-VL 32B.

In all cases, the model receives (i) the action programs or action descriptions for the six positive and six negative examples, and (ii) the corresponding symbolic representation plus rendered image for the query. The task and evaluation protocol are identical to the main experiments (Section~\ref{sec:results}).

Table~\ref{tab:grounded_cg_models} reports per-model performance across Bongard categories for the two symbolic formats, and Table~\ref{tab:grounded_cg_summary} summarizes mean accuracy across the four VLMs, aggregating Human-designed Combined and Novel (HD Comb, HD Novel) into a single HD score. Note that, unlike Table~\ref{tab:main_results}, which averages over 12 models, the statistics here are computed over this smaller matched VLM subset and should be compared only within that subset.

\begin{table}[t]
\centering
\caption{Grounded C--G performance for VLM-capable models. Each entry is accuracy (\%) on Bongard-LOGO categories when symbolic input (Action Description or Action Program) is augmented with the rendered query image. BD = Basic, FF = Free-form, HD Comb/HD Novel = Human-designed shape subsets.}
\label{tab:grounded_cg_models}
\small
\begin{tabular}{l l c c c c}
\toprule
\textbf{Model} & \textbf{Experiment} & \textbf{BD} & \textbf{FF} & \textbf{HD Comb} & \textbf{HD Novel} \\
\midrule
\multirow{2}{*}{Gemma3:12B}
 & Action Desc & 55.0 & 56.2 & 56.8 & 56.0 \\
 & Action Prog & 56.31 & 58.8 & 57.2 & 60.12 \\
\addlinespace
\multirow{2}{*}{Gemma3:27B}
 & Action Desc & 56.2 & 69.2 & 57.6 & 61.0 \\
 & Action Prog & 59.0 & 70.4 & 65.0 & 63.4 \\
\addlinespace
\multirow{2}{*}{LLaVA-Llama3}
 & Action Desc & 49.6 & 50.1 & 50.0 & 50.1 \\
 & Action Prog & 50.2 & 49.0 & 49.2 & 50.2 \\
\addlinespace
\multirow{2}{*}{Qwen2.5VL:32B}
 & Action Desc & 60.4 & 69.2 & 57.8 & 57.4 \\
 & Action Prog & 55.8 & 71.2 & 59.6 & 61.0 \\
\bottomrule
\end{tabular}
\end{table}

For ease of comparison with the main results (Table~\ref{tab:main_results}), we also report mean accuracy and standard deviation across the four VLMs, averaging HD Comb and HD Novel into a single Human-designed score:

\begin{table}[t]
\centering
\caption{Grounded C--G: mean accuracy (\%) $\pm$ standard deviation across VLM-capable models (N=4). HD is the average of HD Comb and HD Novel. These numbers are not directly comparable to the 12-model pooled rows in Table~\ref{tab:main_results}.}
\label{tab:grounded_cg_summary}
\small
\begin{tabular}{l c c c}
\toprule
\textbf{Experiment} & \textbf{BD} & \textbf{FF} & \textbf{HD} \\
\midrule
Action Description & $55.3 \pm 4.5$ & $61.2 \pm 9.6$ & $55.8 \pm 4.0$ \\
Action Program     & $55.3 \pm 3.7$ & $62.4 \pm 10.6$ & $58.2 \pm 6.1$ \\
Baseline           & $57.1 \pm 5.5$ & $62.6 \pm 11.2$ & $57.5 \pm 5.4$ \\
\bottomrule
\end{tabular}
\end{table}

Overall, the grounded setting yields only modest changes relative to the corresponding symbolic-only baselines for this matched VLM subset, and Action Programs retain a small advantage on Free-form and Human-designed categories. This reinforces our main conclusion that symbolic structure, rather than limited visual anchoring, is the primary source of gains on Bongard-LOGO.

\subsection{Randomization Experiments}
\label{app:randomization}

To probe whether high C--G performance reflects genuine rule learning over symbolic structure rather than exploitation of superficial token statistics, we conduct two perturbation experiments in the Action Program setting:

\begin{itemize}
    \item \textbf{Categories Shuffle}: for each Bongard-LOGO problem, we randomly permute the assignment of the 12 support examples to the positive and negative sets while leaving the query program, query image, and gold query label unchanged. This preserves marginal token distributions within a problem but destroys the coherent support-set partition that defines the concept.
    \item \textbf{Test Sequence Shuffle}: for each problem, we randomly shuffle the order of drawing actions in the \emph{query} program at inference time, while leaving the support examples, query image, and gold query label unchanged. This maintains the multiset of action tokens but disrupts the compositional structure of the query.
\end{itemize}

Both perturbations use the same prompts and decoding settings as the main Action Program condition (Section~\ref{sec:experiments}). We report them only on Basic (BD) and Free-form (FF), since the lower unperturbed HD baseline would make the controls less diagnostic there. Table~\ref{tab:shuffle_detailed} reports per-model accuracies on BD and FF, and Table~\ref{tab:shuffle_summary} summarizes mean accuracy across the nine models included in this analysis.

\begin{table}[t]
\centering
\small
\caption{
Performance of C--G (Action Program) under two randomization controls.
\textbf{Categories Shuffle} randomly permutes the assignment of support examples to positive vs.\ negative sets within each problem, while leaving the query and its gold label unchanged.
\textbf{Test Sequence Shuffle} randomly shuffles the order of drawing actions in the query program at inference time, leaving the support examples, query image, and gold label unchanged.
Values show accuracy (\%) on Basic (BD) and Free-form (FF) problems for each model.
}
\label{tab:shuffle_detailed}
\begin{tabular}{l l c c}
\toprule
\textbf{Model} & \textbf{Shuffle Type} & \textbf{BD} & \textbf{FF} \\
\midrule
\multirow{2}{*}{DeepSeek-R1:14B}
 & Categories    & 63.0 & 63.0 \\
 & Test Sequence & 61.6 & 52.2 \\
\addlinespace
\multirow{2}{*}{DeepSeek-R1:32B}
 & Categories    & 58.8 & 73.0 \\
 & Test Sequence & 68.2 & 68.8 \\
\addlinespace
\multirow{2}{*}{DeepSeek-R1:8B}
 & Categories    & 54.3 & 54.0 \\
 & Test Sequence & 54.6 & 51.4 \\
\addlinespace
\multirow{2}{*}{Gemma3:12B}
 & Categories    & 57.6 & 63.2 \\
 & Test Sequence & 64.7 & 54.2 \\
\addlinespace
\multirow{2}{*}{Gemma3:27B}
 & Categories    & 58.6 & 76.6 \\
 & Test Sequence & 65.0 & 55.4 \\
\addlinespace
\multirow{2}{*}{Magistral}
 & Categories    & 64.2 & 70.9 \\
 & Test Sequence & 66.4 & 61.4 \\
\addlinespace
\multirow{2}{*}{Phi-4}
 & Categories    & 61.4 & 81.2 \\
 & Test Sequence & 60.0 & 53.4 \\
\addlinespace
\multirow{2}{*}{Phi-4-Reasoning}
 & Categories    & 57.8 & 86.8 \\
 & Test Sequence & 65.2 & 52.6 \\
\addlinespace
\multirow{2}{*}{Qwen2.5:32B}
 & Categories    & 63.8 & 54.6 \\
 & Test Sequence & 71.6 & 67.8 \\
\bottomrule
\end{tabular}
\end{table}

\begin{table}[t]
\centering
\small
\caption{
Mean accuracy (\%) $\pm$ standard deviation across the nine models in Table~\ref{tab:shuffle_detailed} for each randomization control.
BD = Basic shapes, FF = Free-form shapes.
}
\label{tab:shuffle_summary}
\begin{tabular}{l c c}
\toprule
\textbf{Shuffle Type} & \textbf{BD} & \textbf{FF} \\
\midrule
Categories Shuffle    & $59.9 \pm 3.3$ & $69.3 \pm 11.4$ \\
Test Sequence Shuffle & $64.1 \pm 4.9$ & $57.5 \pm 6.8$ \\
\bottomrule
\end{tabular}
\end{table}

Overall, Categories Shuffle and Test Sequence Shuffle both reduce accuracy relative to the unperturbed Action Program setting, but they need not drive performance exactly to 50\% because the shuffled support sets can still instantiate alternative rules that partially align with the unchanged query. The relevant signal is the relative degradation: Categories Shuffle primarily damages coherent support-set reasoning, whereas Test Sequence Shuffle particularly harms Free-form performance by disrupting the ordered structure of the query. These effects support the interpretation that models rely on the relational and compositional structure of the symbolic programs, especially for Free-form tasks, rather than on simple surface-level token matching.

\subsection{Class Asymmetry in Rule Induction}
\label{app:class_asymmetry}

Beyond overall accuracy, we analyze how models treat positive vs.\ negative examples under the Action Program (AP) interface.
Aggregating across models, we find a pronounced asymmetry: on Basic (BD) problems, accuracy on positive examples is 91.2\% compared to 53.4\% on negatives (a +37.8 point gap), and on Free-form (FF) problems, positives reach 98.5\% vs.\ 61.4\% for negatives (+37.1 points).
A chi-square test of correctness~$\times$~class yields $\chi^2 \approx 1156.3$ for BD and $\chi^2 \approx 1504.2$ for FF ($p \ll 0.001$ in both cases), indicating that correctness is strongly dependent on whether an item belongs to the positive or negative set.
Consistent with this, the pooled predicted labels are skewed toward ``positive'' (e.g., 4{,}799 predicted positives vs.\ 2{,}201 negatives on FF), suggesting that models default to classifying borderline or ambiguous cases as positive rather than negative.

\section{Limitations}
\label{app:limitations}

\subsection{Synthetic Benchmark Limitations}

While Bongard-LOGO provides a controlled environment for hypothesis testing, it is fundamentally limited:
\begin{itemize}
    \item Simple geometric shapes with programmatic rules (not learned from data)
    \item Absence of real-world noise, occlusion, variation, and clutter
    \item Binary classification rather than open-ended reasoning
    \item Formal, unambiguous rules rather than statistical regularities
\end{itemize}

Results on this benchmark should not be interpreted as evidence of general visual reasoning capability. Evaluation on natural-image Bongard datasets (e.g., Bongard-OpenWorld with VLM–LLM interfaces) and other diverse benchmarks is essential for establishing broader generalization.

\subsection{Ground-Truth Privilege and Practical Applicability}

Our approach assumes access to perfect symbolic specifications—the ground-truth action programs used to generate each image. This represents a significant privilege that is unavailable in most real-world scenarios:
\begin{itemize}
    \item Most domains lack formal generative specifications
    \item Obtaining symbolic descriptions requires either domain-specific engineering, learned perceptual systems, or human annotation
\end{itemize}

We caution against over-interpreting the success of the C-G approach as evidence that human-level visual reasoning is imminent. Rather, it illustrates what becomes possible when perception is perfectly decoupled from reasoning.

\subsection{Model Coverage Gaps}

Our evaluation covered recent reasoning-capable models but excluded:
\begin{itemize}
    \item Vision-language models in their symbolic reasoning mode (architectural constraints limited multi-image symbolic reasoning)
    \item Older and specialized models (e.g., earlier GPT/Claude variants, mathematical reasoners) were not evaluated systematically.
    \item Models from certain organizations (OpenAI, Anthropic APIs were not used)
    \item Smaller models (\textless 7B parameters) were not systematically evaluated
\end{itemize}

Results on a broader model portfolio would strengthen generality claims and establish baseline expectations for future work.

\subsection{Environmental and Computational Impact}

This research required:
\begin{itemize}
    \item €1000 in API costs to run frontier models (financial barrier to reproducibility)
    \item 2000+ GPU-hours (computational barrier)
    \item Significant energy consumption (environmental impact)
\end{itemize}

These barriers limit reproducibility and raise equity concerns. 

\subsection{Incomplete Analysis}

Several aspects of this work merit deeper investigation:
\begin{itemize}
    \item Human-designed performance ($\approx$60--66\%) remains substantially lower than the BD and FF regimes despite perfect symbolic input.
    \item Robustness to symbolic input corruption (e.g., noise in programs or descriptions) has not been evaluated.
\end{itemize}

Further work is needed to analyze failure modes in output parsing and to characterize computational efficiency and scaling behavior.

\end{document}

%% file: per_model_condition_plot.tex
\begin{tikzpicture}
\begin{groupplot}[
group style={group size=3 by 1, horizontal sep=1.2cm},
width=0.31\textwidth,
height=0.24\textheight,
ymin=-12, ymax=14,
xmin=-0.35, xmax=2.35,
xtick={0,1,2},
xticklabels={AD$-$Base,AP$-$Base,+C on AP},
tick label style={font=\scriptsize},
title style={font=\small},
ylabel style={font=\small},
grid=both,
major grid style={draw=gray!20},
minor grid style={draw=gray!10},
minor y tick num=1,
legend style={draw=none, font=\scriptsize, /tikz/every even column/.append style={column sep=0.3cm}},
]

\nextgroupplot[title={Free-form (FF)}, ylabel={Accuracy Change (pp)}]
\addplot[dashed, black!55] coordinates {(-0.35,0) (2.35,0)};
\addplot[only marks, mark=*, mark size=1.4pt, draw=gray!70, fill=gray!70] coordinates {
(-0.180,11.6) (-0.145,-0.6) (-0.110,8.2) (-0.075,-2.0) (-0.040,-1.8) (-0.005,2.6) (0.030,-6.8) (0.065,-9.2) (0.100,5.0) (0.135,7.2) (0.170,4.2) (0.205,5.0)
(0.820,4.8) (0.855,3.2) (0.890,4.0) (0.925,-1.2) (0.960,0.4) (0.995,-1.8) (1.030,-7.2) (1.065,-0.8) (1.100,0.4) (1.135,6.8) (1.170,3.4) (1.205,-3.0)
(1.820,-0.2) (1.855,-2.2) (1.890,2.6) (1.925,1.4) (1.960,4.2) (1.995,1.2) (2.030,2.2) (2.065,-2.0) (2.100,2.0) (2.135,0.6) (2.170,1.8) (2.205,3.4)
};
\addplot[only marks, mark=diamond*, mark size=3.1pt, draw=black, fill=black] coordinates {(0,1.95) (1,0.75) (2,1.25)};
\node[font=\scriptsize, anchor=south, fill=white, inner sep=1pt] at (axis cs:0,1.95) {+2.0};
\node[font=\scriptsize, anchor=south, fill=white, inner sep=1pt] at (axis cs:1,0.75) {+0.8};
\node[font=\scriptsize, anchor=south, fill=white, inner sep=1pt] at (axis cs:2,1.25) {+1.3};
\addlegendentry{One model}
\addlegendentry{Mean delta}

\nextgroupplot[title={Basic (BD)}]
\addplot[dashed, black!55] coordinates {(-0.35,0) (2.35,0)};
\addplot[only marks, mark=*, mark size=1.4pt, draw=gray!70, fill=gray!70] coordinates {
(-0.180,12.6) (-0.145,-1.4) (-0.110,-1.6) (-0.075,-1.54) (-0.040,-2.0) (-0.005,3.2) (0.030,5.0) (0.065,9.2) (0.100,2.4) (0.135,8.4) (0.170,9.8) (0.205,7.2)
(0.820,4.52) (0.855,-3.0) (0.890,-1.0) (0.925,-2.6) (0.960,-0.6) (0.995,-2.0) (1.030,0.6) (1.065,2.4) (1.100,2.2) (1.135,6.6) (1.170,7.2) (1.205,-1.0)
(1.820,-1.52) (1.855,0.8) (1.890,2.2) (1.925,-1.2) (1.960,2.4) (1.995,2.2) (2.030,0.8) (2.065,3.0) (2.100,0.8) (2.135,0.4) (2.170,3.0) (2.205,1.2)
};
\addplot[only marks, mark=diamond*, mark size=3.1pt, draw=black, fill=black] coordinates {(0,4.27) (1,1.11) (2,1.17)};
\node[font=\scriptsize, anchor=south, fill=white, inner sep=1pt] at (axis cs:0,4.27) {+4.3};
\node[font=\scriptsize, anchor=south, fill=white, inner sep=1pt] at (axis cs:1,1.11) {+1.1};
\node[font=\scriptsize, anchor=south, fill=white, inner sep=1pt] at (axis cs:2,1.17) {+1.2};

\nextgroupplot[title={Human-designed (HD)}]
\addplot[dashed, black!55] coordinates {(-0.35,0) (2.35,0)};
\addplot[only marks, mark=*, mark size=1.4pt, draw=gray!70, fill=gray!70] coordinates {
(-0.180,4.5) (-0.145,-1.6) (-0.110,2.1) (-0.075,-0.3) (-0.040,-4.9) (-0.005,0.0) (0.030,-4.8) (0.065,-2.0) (0.100,-1.3) (0.135,1.0) (0.170,0.0) (0.205,0.3)
(0.820,3.46) (0.855,-1.2) (0.890,4.0) (0.925,3.0) (0.960,1.4) (0.995,-2.7) (1.030,1.0) (1.065,4.0) (1.100,0.5) (1.135,2.2) (1.170,0.4) (1.205,-0.2)
(1.820,-0.06) (1.855,0.2) (1.890,-1.7) (1.925,-1.8) (1.960,-1.9) (1.995,3.4) (2.030,0.2) (2.065,0.1) (2.100,1.7) (2.135,1.1) (2.170,-0.3) (2.205,0.9)
};
\addplot[only marks, mark=diamond*, mark size=3.1pt, draw=black, fill=black] coordinates {(0,-0.58) (1,1.32) (2,0.15)};
\node[font=\scriptsize, anchor=north, fill=white, inner sep=1pt] at (axis cs:0,-0.58) {-0.6};
\node[font=\scriptsize, anchor=south, fill=white, inner sep=1pt] at (axis cs:1,1.32) {+1.3};
\node[font=\scriptsize, anchor=south, fill=white, inner sep=1pt] at (axis cs:2,0.15) {+0.2};

\end{groupplot}
\end{tikzpicture}